\documentclass[final]{nesy2026} 
\usepackage{booktabs}

\title[Ontology-Grounded Project Memory]{Ontology-Grounded Project Memory for Coding Agents}

\clearauthor{\Name{James Adam} \Email{james@trivyn.io}\\
\addr Trivyn}

\begin{document}

\maketitle

\begin{abstract}
Coding agents have become the primary means of generating new code in many
software projects, and the resulting velocity of changes makes keeping track of the reasons
behind those changes challenging. This paper introduces MOOSEDev, a system designed to give
coding agents structured, ontology-grounded project memory. The system captures architectural decisions, lessons,
constraints, and rationales in a knowledge graph exposed to agents via a Model Context Protocol (MCP) interface.
Records carry lifecycle status, provenance, and supersession links, queryable via MOOSE,
a proprietary neurosymbolic engine that treats the symbolic layer as the primary reasoning substrate. We compared MOOSEDev against a production vector-memory tool
on a neutral public corpus of 835 typed records. MOOSEDev returned the expected answer set
essentially in full (0.98--1.00) on supersession, set-completeness, and negation questions,
whereas the baseline's top-$k$ retrieval surfaced between 6\% and 27\%.
Conversely, relevance recall and token cost were largely equivalent between the two systems.
We also describe a temporal commit-history bootstrap of our own codebase, a pre-registered live trial, and lessons learned.
\end{abstract}

\section{Introduction}
\label{sec:intro}

Software teams increasingly use coding agents for the majority of new code, while keeping
architecture and design under the purview of human engineers; teams working this way must
continually remind their agents of design decisions and intervene when they go off track.
The deeper cost is being called ``comprehension debt,'' the gradual loss of understanding of why a
codebase is the way it is. Source code is the manifestation of what was decided;
the reasoning, theory, and rationale behind it live elsewhere \citep{naur1985}.

Notes files, Markdown specifications, and retrieval-augmented memory
\citep{lewis2020,chhikara2025mem0} help, but persistent issues remain. Agents need to know what
kind of knowledge a note represents (decision, constraint, rationale, lesson, anti-pattern);
whether it is current or historical; and how
records relate: this decision supersedes that one.
Those distinctions are fundamentally ontological. A vector search can find nearby words; it does
not know what a record \emph{is} or what lifecycle role it plays. At that point project memory
stops being a note-keeping problem and becomes a modeling problem.
Knowledge graphs over source code itself are well studied, including extracted graphs of classes and
usage \citep{abdelaziz2021} as well as API-evolution graphs used to drive code generation
\citep{kang2026kcoevo}.  While those solutions model the code, MOOSEDev models the knowledge
\emph{about} it, a layer that work leaves implicit.

This paper is a case study of treating coding agent memory as an ontology problem. 
We present MOOSEDev \citep{adam2026moosedev}, a deployed neurosymbolic memory
system (Section~\ref{sec:system}) with a head-to-head evaluation against a production vector-memory
tool under a validated strict judge (Section~\ref{sec:eval}), plus deployment workflows,
lessons learned, and adoption barriers (Section~\ref{sec:deploy}).

\section{System: Typed Records on a Neurosymbolic Engine}
\label{sec:system}

MOOSEDev records architectural decisions \citep{nygard2011}, lessons, constraints, rationales, and anti-patterns in
a project knowledge graph, grounded in two small ontologies:
software-engineering for the structural vocabulary of a codebase, and software-architecture for the knowledge
\emph{about} that structure. Both are OWL ontologies with companion SHACL shapes
\citep{w3c2012owl2,knublauch2017shacl}, generated interactively in Trivyn, our
ontology-engineering workbench, refined manually, and reasoned over at runtime by the MOOSE
engine; both are deliberately small, nine and eleven classes respectively, 51 properties in
total. The relationships are crucial: a record can carry a rationale, supersede another record, affect
a component, and bear lifecycle status, author, and timestamp; these edges form a traversable
network.

Structure provides the agent with typed capture, validation against SHACL shapes,
queryability (e.g. ``accepted decisions with no recorded rationale''), lifecycle and
provenance tracking, and the ability to align new concepts.

The underlying neurosymbolic engine \citep{garcez2023} treats the LLM as an unreliable, albeit useful sensor.
Keyword and structural matching, ontology traversal, deterministic evidence fusion, validation,
and execution traces are symbolic; the model is consulted only at narrow, declared points, so it can be small (8--32B).
When an agent asks for current guidance, the engine traverses the graph to find relevant records, 
follows typed relationships, filters superseded records, and deterministically ranks the remaining 
evidence into context; the LLM only interprets the question and
synthesizes the answer, and every step is logged in an execution trace.

MOOSEDev exposes this capability over MCP \citep{anthropic2024mcp} as four main tool groups: typed capture, reading
(context retrieval, natural-language query, SPARQL \citep{harris2013sparql}), lifecycle, and
integrity.

\section{Evaluation: Structured vs Vector Memory}
\label{sec:eval}

\paragraph{Test setup.}
Does an ontology-grounded knowledge graph justify its complexity against free-text and
vector memory for a coding agent?\footnote{The benchmark harness, CodeGraph corpus export, and public-corpus run transcripts
are at \url{https://github.com/Trivyn/moosedev} (\texttt{bench/}); private-corpus
transcripts are withheld, with aggregates in the repository's evaluation summary.} Our benchmark compares five
memory and retrieval conditions. B0 is the
coding agent with no memory (i.e., ``vibe coding mode''): the floor. B1-notes provides the agent
with a corpus of real documentation as flat files. B1-mem0 is mem0 \citep{chhikara2025mem0}, a contemporary
vector-memory tool ingesting the same documentation with its own native capture method. B1-rag is a basic
BM25 \citep{robertson2009} tool over the graph's own content flattened into text, testing
whether the content alone, without structure, carries the result. B2 is our structured
project memory graph. The test corpus is the documentation of CodeGraph, a third-party open-source developer tool,
captured as 835 typed records in our graph and 553 mem0 memories, verified faithful.
A single CLI agent configuration and model (GPT-5.x family) was used for every run; we
pre-registered the verdicts, authored ground truth from primary sources or SPARQL-derived
sets, and graded with a strict LLM judge (GPT-5.4-mini) that allowed paraphrase but rejected
answers on the correct topic with the wrong facts, validated by reproducing the graph
condition's strict-match score.\footnote{Mean of three logged judging passes, per-item
verdicts released with the artifact; an earlier unlogged pass deviated at most 0.12 per
cell, in both directions.}
Every run is an immutable transcript, re-gradable without
rerunning the agent.

\begin{table}[htbp]
\floatconts
  {tab:capability}
  {\caption{Answer score (0 to 1) by task class, CodeGraph corpus, validated strict judge.}}
  {\begin{tabular}{lccc}
  \toprule
  \bfseries Task class & \bfseries B2 (typed graph) & \bfseries B1-mem0 & \bfseries B1-notes (docs search)\\
  \midrule
  Set completeness & 1.00 & 0.18 & 0.08\\
  Negation (absence) & 0.98 & 0.06 & 0.00\\
  Supersession traversal & 0.98 & 0.27 & 0.12\\
  \bottomrule
  \end{tabular}}
\end{table}

\paragraph{Capability differences.}
The results in Table~\ref{tab:capability} show a qualitative difference on tasks that require
structural reasoning. (B0 scored zero on these classes and is omitted; B1-rag appears below
via the F1 comparison.) It is not caused by incomplete ingestion: B1-mem0 contains the relevant
facts, but its retrieval returns a limited ranked slice, so it cannot enumerate a complete
set, establish an absence, or traverse relationships. The structured advantage
increases with the size and complexity of the required answer (for a two-item set the
systems are equivalent), consistently across corpora. Against B1-rag, set-overlap F1
was 0.94 versus 0.25 on the private corpus and 0.90 versus 0.34 on the neutral public one.
The capability does cost tokens, however: 78k to 167k for a 122-item completeness set via
SPARQL, compared to roughly 35k for targeted retrieval. The baseline systems, by contrast, could not
produce the complete answer at any cost.  This demonstrates the idea that memory is a modeling
problem: where a question is structural (e.g., completeness, absence, traversal), the 
advantage over basic retrieval is categorical.

\paragraph{Parity on retrieval, currency by construction.}
Testing only tasks where structured retrieval is required would be a bit of a strawman, so
the matrix also tests where vector memory is expected to excel, and there the conditions tie.
On simple relevance the graph reached coverage 0.82 at roughly 35k agent tokens, whereas mem0 scored
between 0.67 and 0.90 at roughly 40k. We had expected the structured advantage to
compound as the store grew, but it did not: a scale study from 50 to 634 records found the
graph ahead at every size (hit@5 0.84 versus 0.60 at the largest) at a constant offset; the
slope's confidence interval spans zero.
Retrieval precision is thus not the differentiator, and on its own it would likely not justify
the system. Currency is different: across four real-world reversal pairs, four models, and two
delivery regimes, the graph served the current answer in all 40 trials, since superseded
records are excluded from current-guidance retrieval by construction. On the reversal tested,
B1-mem0 tied the graph at 100\% where a cold agent scored 0.00; the free-text B1-rag condition,
on the one pair where the superseded record outranks the current one and with that content
pushed into the prompt, was current in only 1 of 13 trials (8\%).

\section{Deployment, Bootstrap, and Lessons}
\label{sec:deploy}

\paragraph{Bootstrapping existing repositories.}
A graph built from only the current state of a repository would be flat, missing the
historical relationships, so our bootstrap workflow walks repository history commit by
commit, extracting typed records with historical timestamps. Over our own
repository, it recovered seven real supersession chains. The ontologies are domain-level
(e.g., software architecture concepts) rather than project-specific, so a new codebase 
needs no custom schema work, only the instance graph built during bootstrap.

\paragraph{A pre-registered live trial.}
The claim that motivated this project, that structured memory reduces comprehension debt
over months of real work, cannot be tested by a point-in-time benchmark, so we are testing
the system in daily use on two production codebases at our company. 
We established sixteen fixed reference points for context recovery at the start of testing,
each targeting a reversal whose earlier approach is deleted from
the current tree, so the answer cannot be reconstructed from the code alone. Gold-standard answers were authored from primary
sources, never from the graph, so the graph cannot grade itself. A blind local judge scores a
monthly graph-versus-cold gap against a recorded month-zero baseline, with a pre-set failure
condition: fewer than four record-citing assists per project per month triggers the
system's retirement. The registered readout is the monthly trend; in the trial's first
twenty-one days the graph provided 71 unprompted assists citing relevant records, while
logging 38 misses (an unhelpful recall or knowledge it should have carried), 35 of 38 in
one project bootstrapped from a thinner commit history.

\paragraph{Lessons for practitioners.}
Measuring an LLM-based memory system misleads in predictable ways. Our evaluation nearly
shipped three wrong conclusions: a 20-point win for our own system that was a grading
artifact; an apparent relevance loss caused by a configuration bug: the memory server was down
and agents silently grepped the raw store; and a misleading commit message that poisoned
both a gold answer and a bootstrapped rationale. The rules that caught them generalize beyond
benchmarks: immutable transcripts with free re-grades, a judge validated against a condition
with a known score, runs that verify the memory tool actually fired, and ground truth only
from primary sources. Each time a plausible-but-wrong conclusion almost stuck, an auditable,
re-checkable process caught it: exactly the failure mode (confident, fluent, wrong) this
class of system exists to reduce.

\paragraph{Barriers to adoption.}
The capability win depends on correct capture: flat capture degrades the graph to free-text
parity, and that discipline imposes friction that a vector store does not. Alignment tooling,
assisted capture, and the conclusion's push direction all lower that cost. The design
also leans on host tool use: MCP is passive; the agent decides when to call, which favors
higher-tier models; we evaluated one agent family (GPT-5.x).

\section{Conclusion}
\label{sec:conclusion}

Against a common vector-memory tool, ontology-grounded memory (given effective capture) ties where
retrieval is strong, and is clearly better on completeness, absence, and
supersession. Small local models with limited context windows should benefit most, so the
natural next step is to act inside the loop: pushing entity-exact records when code is
touched and gating edits on recorded constraints.

\bibliography{refs}

\begin{thebibliography}{13}
\providecommand{\natexlab}[1]{#1}
\providecommand{\url}[1]{\texttt{#1}}
\expandafter\ifx\csname urlstyle\endcsname\relax
  \providecommand{\doi}[1]{doi: #1}\else
  \providecommand{\doi}{doi: \begingroup \urlstyle{rm}\Url}\fi

\bibitem[Abdelaziz et~al.(2021)Abdelaziz, Dolby, McCusker, and
  Srinivas]{abdelaziz2021}
Ibrahim Abdelaziz, Julian Dolby, James~P. McCusker, and Kavitha Srinivas.
\newblock A toolkit for generating code knowledge graphs.
\newblock In \emph{Proceedings of the 11th Knowledge Capture Conference
  ({K-CAP})}, 2021.
\newblock \url{https://doi.org/10.1145/3460210.3493578}.

\bibitem[Adam(2026)]{adam2026moosedev}
James Adam.
\newblock {MOOSEDev}: A practical application of ontologies and the {MOOSE}
  engine.
\newblock Blog post, 2026.
\newblock \url{https://trivyn.io/blog/introducing-moosedev}.

\bibitem[{Anthropic}(2024)]{anthropic2024mcp}
{Anthropic}.
\newblock {Model Context Protocol}, 2024.
\newblock \url{https://modelcontextprotocol.io}.

\bibitem[Chhikara et~al.(2025)Chhikara, Khant, Aryan, Singh, and
  Yadav]{chhikara2025mem0}
Prateek Chhikara, Dev Khant, Saket Aryan, Taranjeet Singh, and Deshraj Yadav.
\newblock Mem0: Building production-ready {AI} agents with scalable long-term
  memory, 2025.
\newblock arXiv:2504.19413, \url{https://arxiv.org/abs/2504.19413}.

\bibitem[d'Avila Garcez and Lamb(2023)]{garcez2023}
Artur d'Avila Garcez and Lu{\'i}s~C. Lamb.
\newblock Neurosymbolic {AI}: The 3rd wave.
\newblock \emph{Artificial Intelligence Review}, 56\penalty0 (11):\penalty0
  12387--12406, 2023.
\newblock arXiv:2012.05876. \url{https://doi.org/10.1007/s10462-023-10448-w}.

\bibitem[Harris and Seaborne(2013)]{harris2013sparql}
Steve Harris and Andy Seaborne.
\newblock {SPARQL 1.1 Query Language}, 2013.
\newblock W3C Recommendation, 21 March 2013,
  \url{https://www.w3.org/TR/sparql11-query/}.

\bibitem[Kang et~al.(2026)Kang, Lu, Jiang, Liu, Zhang, Jiang, Sun, Wu, and
  Qi]{kang2026kcoevo}
Jiazhen Kang, Yuchen Lu, Chen Jiang, Jinrui Liu, Tianhao Zhang, Bo~Jiang,
  Ningyuan Sun, Tongtong Wu, and Guilin Qi.
\newblock {KCoEvo}: A knowledge graph augmented framework for evolutionary code
  generation, 2026.
\newblock arXiv:2603.07581, \url{https://arxiv.org/abs/2603.07581}.

\bibitem[Knublauch and Kontokostas(2017)]{knublauch2017shacl}
Holger Knublauch and Dimitris Kontokostas.
\newblock {Shapes Constraint Language (SHACL)}, 2017.
\newblock W3C Recommendation, 20 July 2017, \url{https://www.w3.org/TR/shacl/}.

\bibitem[Lewis et~al.(2020)Lewis, Perez, Piktus, Petroni, Karpukhin, Goyal,
  K{\"u}ttler, Lewis, Yih, Rockt{\"a}schel, Riedel, and Kiela]{lewis2020}
Patrick Lewis, Ethan Perez, Aleksandra Piktus, Fabio Petroni, Vladimir
  Karpukhin, Naman Goyal, Heinrich K{\"u}ttler, Mike Lewis, Wen{-}tau Yih, Tim
  Rockt{\"a}schel, Sebastian Riedel, and Douwe Kiela.
\newblock Retrieval-augmented generation for knowledge-intensive {NLP} tasks.
\newblock In \emph{Advances in Neural Information Processing Systems 33}, pages
  9459--9474, 2020.
\newblock \url{https://arxiv.org/abs/2005.11401}.

\bibitem[Naur(1985)]{naur1985}
Peter Naur.
\newblock Programming as theory building.
\newblock \emph{Microprocessing and Microprogramming}, 15\penalty0
  (5):\penalty0 253--261, 1985.
\newblock \url{https://doi.org/10.1016/0165-6074(85)90032-8}.

\bibitem[Nygard(2011)]{nygard2011}
Michael Nygard.
\newblock Documenting architecture decisions.
\newblock Blog post, 2011.
\newblock
  \url{https://cognitect.com/blog/2011/11/15/documenting-architecture-decisions}.

\bibitem[Robertson and Zaragoza(2009)]{robertson2009}
Stephen Robertson and Hugo Zaragoza.
\newblock The probabilistic relevance framework: {BM25} and beyond.
\newblock \emph{Foundations and Trends in Information Retrieval}, 3\penalty0
  (4):\penalty0 333--389, 2009.
\newblock \url{https://doi.org/10.1561/1500000019}.

\bibitem[{W3C OWL Working Group}(2012)]{w3c2012owl2}
{W3C OWL Working Group}.
\newblock {OWL 2 Web Ontology Language Document Overview (Second Edition)},
  2012.
\newblock W3C Recommendation, 11 December 2012,
  \url{https://www.w3.org/TR/owl2-overview/}.

\end{thebibliography}

\end{document}